\documentstyle[12pt]{article}
\newcommand{\mf}{\tt} 
\newcommand{\type}{\em}
\newcommand{\nterm}[1]{\mbox{\em #1}}
\newcommand{\tag}[1]{\fbox{\footnotesize #1}}
\newcommand{\unif}{\sqcup}
\newcommand{\subsumes}{\sqsubseteq}
\newcommand{\get}{\mbox{$\leftarrow\;$}}
\newcommand{\edgedot}{\bullet}

\newenvironment{tfs}[1]{\left[ \begin{array}{ll}{\mbox{\bf
#1}}\\}{\end{array} \right]}

\newcommand{\Tabs}{xxxx\= xxxx\= xxxx\= xxxx\= xxxx\= xxxx\= xxxx\= xxxx\= xxxx
\= xxxx\= xxxx\= xxxx\= \kill}
\newenvironment{program}[3]{\begin{figure*}[hbtp]
                         \begin{center}
                         \fbox{\footnotesize\mf
                         \begin{minipage}{\textwidth}
                         \begin{tabbing}
                         \Tabs
                         #3
                         \end{tabbing}
                         \end{minipage}
                         }
                         \end{center}
                         \caption{#2}
                         \label{#1}
                         \end{figure*}}{}

\begin{document}
\title{Abstract Machine for Typed Feature Structures}
\author{Shuly Wintner \and Nissim Francez}
\date{Computer Science\\
	Technion, Israel Institute of Technology\\
	32000 Haifa, Israel\\
	{\tt \{shuly,francez\}@cs.technion.ac.il}}
\maketitle
\begin{abstract}
This paper describes an abstract machine for linguistic formalisms
that are based on typed feature structures, such as HPSG.  The core
design of the abstract machine is given in detail, including the
compilation process from a high-level language to the abstract machine
language and the implementation of the abstract instructions.  The
machine's engine supports the unification of typed, possibly cyclic,
feature structures.  A separate module deals with control structures
and instructions to accommodate parsing for phrase structure grammars.
We treat the linguistic formalism as a high-level declarative
programming language, applying methods that were proved useful in
computer science to the study of natural languages: a grammar
specified using the formalism is endowed with an operational
semantics. \\
{\bf Topics:} Grammar formalisms, feature structures, Parsing,
Compilation, WAM.
\end{abstract}
\thispagestyle{empty}

\section{Introduction}
Typed feature structures (TFSs) serve as a means for the specification
of linguistic information in current linguistic formalisms such as
HPSG (\cite{hpsg1,hpsg2}) or Categorial Grammar (\cite{cg}).
Generalizing first-order terms (FOTs), TFSs are also used to specify
logic programs and constraint systems in LOGIN (\cite{login}), LIFE
(\cite{life}), ALE (\cite{carpolfran,ale}), TFS (\cite{tfs}) and
others.  General frameworks that are completely independent of any
linguistic theory can be used to specify grammars for natural
languages. Indeed, most of the above mentioned languages were used for
specifying HPSG grammars.

Linguistic formalisms (in particular, HPSG) use TFSs as the basic
blocks for representing linguistic data: lexical items, phrases and
rules. Usually, no mechanism for manipulating TFSs (e.g., parsing
algorithm) is specified.  Current approaches for processing HPSG
grammars either translate the grammar to Prolog (e.g.,
\cite{penn,profit}) or specify it as a general constraint system.
Using general solvers for a specific application, namely parsing,
results in disappointing performance. Clearly, efficient processing
calls for a different method.

Research in the semantics of programming language has undergone much
progress in recent years. At the same time, linguistic theories have
become more formal and grammars for natural languages are nowadays
specified with rigor, resembling computer programs. The interaction of
computer science and linguistics enables the use of techniques and
results of the former to be applied to the latter.

We present an approach for processing TFSs that guarantees both an
explicit definition and high efficiency. Our main aim is to provide an
operational semantics for TFS-based linguistic formalisms, especially
HPSG. We adopt an abstract machine approach for the compilation of
grammars, in a formalism that is a subset of ALE.  Such approaches
were used for processing procedural and functional languages, but they
gained much popularity for logic programming languages since the
introduction of the Warren Abstract Machine (WAM -- see
\cite{wam}). Most current implementations of Prolog, as well as of
other logic languages, are based on abstract machines.  The
incorporation of such techniques usually leads to very efficient
compilers in terms of both space and time requirements.  The abstract
machine is composed of data structures and a set of instructions,
augmented by a compiler from the TFS formalism to the abstract
instructions. The effect of each instruction is defined using a
low-level language that can be executed on ordinary
hardware. Recently, a similar approach was applied to LIFE
(\cite{prl7}), which is a general purpose logic programming language;
however, due to differences in the motivation and in the formalisms,
our machine is much different. Moreover, the LIFE machine is limited
to term unification, whereas our machine includes a control module
that enables manipulation of whole grammars.

The abstract machine ensures that a grammars specified using our
system are endowed with well defined meaning. It enables, for example,
to formally verify the correctness of a compiler for HPSG, given an
independent definition.  The design of such an abstract architecture
must be careful enough to compromise two, usually conflicting,
requirements: the closer the machine is to common architectures, the
harder it is to develop compilers for it; on the other hand, if such a
machine is too complex, then while a compiler for it is easier to
produce, it becomes more complicated to execute its language on normal
architectures.

The next section sketches the framework for which our machine is
designed and defines some basic notions. Section~\ref{m0} describes
the abstract machine core design along with the compilation scheme. In
section~\ref{parsing} control structures are added to enable parsing.
A conclusion and plans for further research are given in
section~\ref{conclusion}.  Due to lack of space, the description is
rather general. Refer to \cite{shuly:tr-lcl-94-8} for more details.

\section{The Framework}
\label{framework}
\subsection{Fundamental Notions}
We briefly review the basic notions we use (thoroughly described in
\cite{carp92,shuly:tr-lcl-94-8}).
An HPSG grammar consists of a type specification and grammar rules
(including principles and lexical rules).  The basic entity of HPSG is
the (typed) {\bf feature structure} (TFS), which is a connected, directed,
labeled, possibly cyclic, finite graph, whose nodes are decorated with
{\bf types} and whose edges are labeled by {\bf features}.  A TFS is
{\bf reentrant} if it contains two different paths that lead to the
same node.  The types are ordered according to an {\bf inheritance
hierarchy} where higher types inherit features from their super-types.

Many different formalizations of TFS systems exist; we basically
follow the definitions of (\cite{ale,carp92}).  The set of types
includes both $\bot$, the least type, and $\top$, the greatest
one. Types are ordered by {\bf subsumption} ($\subsumes$) according to
their information content, not set inclusion of their
denotation. Hence, $\bot$ is the most general type, subsuming every
other, and $\top$ is the contradictory type, subsumed by every other.

The inheritance hierarchy is required to be bounded complete: every
set of consistent types $t_1,\ldots, t_n$ must have a unique least
upper bound $t_1 \unif \cdots \unif t_n \neq \top$.  Every partial
order can be naturally extended to a bounded complete one. The
appropriateness function $Approp(t,f)$ is required to be monotone and
to comply with the feature introduction condition.\footnote{This
condition states that every feature is introduced by some least type
and is appropriate for all the types it subsumes.}  However, we allow
appropriateness specifications to contain loops.

The basic operation performed on TFSs is {\bf unification}
($\unif$). There are various definitions for TFS unification, and we
base our unification algorithm on the definition given in
\cite{carp92}. Two TFSs $A$ and $B$ are {\bf inconsistent} if their
unification results in {\bf failure}, denoted by $A \unif B = \top$.

The TFSs with which we deal are required to be totally
well-typed,\footnote{A TFS is totally well-typed if it contains all
and only the features that are appropriate for its type, and each
feature bears an appropriate value. This requirement will be slightly
relaxed; see section~\ref{optimization}.} for more efficient
processing. This might be problematic for the users who may prefer to
specify only partial information about linguistic entities. Therefore,
some description language must be provided, allowing partial
descriptions from which totally well-typed feature structures can be
automatically deduced.  As there are efficient algorithms to deduce
structures from their descriptions, we prefer not to commit ourselves
to one description language. We define our system over explicit
representations of TFS, as will be clear from
section~\ref{representation}.

\subsection{Type Specification}
A program (or a grammar) contains a type specification, consisting of
a type hierarchy and an appropriateness specification. We adopt ALE's
format (\cite{ale}) for this specification: it is a sequence of
statements of the form:
\begin{center}
$t$ {\mf sub} $[t_1,t_2,\ldots,t_n]$ {\mf intro} $[f_1:r_1,\ldots,f_m:r_m]$.
\end{center}
where ${t,t_1,\ldots,t_n,r_1,\ldots,r_m}$ are types,
$f_1,\ldots,f_m\;$ are features and $n,m \ge 0$.  This statement,
which is said to {\bf characterize} $t$, means that ${t_1,\ldots,t_n}$
are (immediate) subtypes of ${t}$ (i.e., for every $i, 1 \le i \le n,
{t} \subsumes {t_i}$), and that ${t}$ has the features
$f_1,\ldots,f_m$ appropriate for it. Moreover, these features are {\bf
introduced} by $t$, i.e., they are not appropriate for any type $t'$
such that $t' \sqsubset t$. Finally, the statement specifies that
$Approp(t,f_i) = r_i$ for every $i$.  Each type (except $\top$ and
$\bot$) must be characterized by exactly one statement. The {\bf
arity} of a type $t$, $Ar(t)$, is the number of features appropriate for it.

The full subsumption relation is the reflexive transitive closure of
the immediate relation determined by the characterization
statements. If this relation is not a bounded complete partial order,
the specification is rendered invalid. The same is true in case it is
not an appropriateness specification.

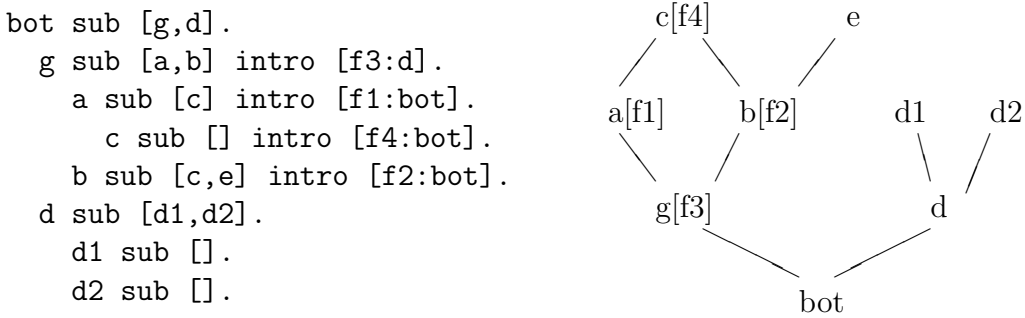
\begin{figure}[hbt]
\begin{minipage}{7.5cm}
\begin{verbatim}
bot sub [g,d].
  g sub [a,b] intro [f3:d].
    a sub [c] intro [f1:bot].
      c sub [] intro [f4:bot].
    b sub [c,e] intro [f2:bot].
  d sub [d1,d2].
    d1 sub [].
    d2 sub [].
\end{verbatim}
\end{minipage} \hfill
\begin{minipage}{7.5cm}
\setlength{\unitlength}{0.012500in}%
\begin{picture}(160,131)(280,440)
\put(320,475){\line( 2,-1){ 40}}
\put(415,475){\line(-2,-1){ 40}}
\put(285,515){\line( 3,-4){ 15}}
\put(335,515){\line(-1,-2){ 10}}
\put(300,555){\line(-3,-4){ 15}}
\put(320,555){\line( 3,-4){ 15}}
\put(375,555){\line(-3,-4){ 15}}
\put(410,515){\line( 1,-4){  5}}
\put(440,515){\line(-2,-5){ 10}}
\put(360,440){\makebox(0,0)[lb]{\smash{\mf}bot}}
\put(280,520){\makebox(0,0)[lb]{\smash{\mf}a[f1]}}
\put(335,520){\makebox(0,0)[lb]{\smash{\mf}b[f2]}}
\put(300,560){\makebox(0,0)[lb]{\smash{\mf}c[f4]}}
\put(380,560){\makebox(0,0)[lb]{\smash{\mf}e}}
\put(400,520){\makebox(0,0)[lb]{\smash{\mf}d1}}
\put(440,520){\makebox(0,0)[lb]{\smash{\mf}d2}}
\put(415,480){\makebox(0,0)[lb]{\smash{\mf}d}}
\put(300,480){\makebox(0,0)[lb]{\smash{\mf}g[f3]}}
\end{picture}
\end{minipage}
\caption{An example type hierarchy}
\label{hier}
\end{figure}

We use the type hierarchy in figure~\ref{hier} as a running example, where
\verb+bot+ stands for $\bot$. The type $\top$
is systematically omitted from type specifications.

\subsection{Representation of Feature Structures}
\label{representation}
\begin{sloppypar}
The most convenient graphical representation of TFSs is
attribute-value matrices (AVMs). However, to represent a (totally
well-typed) feature structure linearly we use an FOT-like notation,
based upon A\"{\i}t-Kaci's $\psi$-terms (\cite{login,osf}), where the
type plays a similar role to that of a function symbol and the
features are listed in a fixed order. Reentrancy is implied by
attaching identical tags to reentrant TFSs. A term is {\bf normal} if
all its types are tagged, and if the same tag appears more than once,
then only its first occurrence carries information.
See~\cite{shuly:tr-lcl-94-8} for the details.
\end{sloppypar}

Total well-typedness implies that the names of the features in a TFS can
be coded by their position in the argument list of a type, and thus
feature-names are omitted from the linear representation.  Assuming
that the feature names are ordered alphabetically, the linear
representation of an example TFS is given in figure~\ref{fs-b}.
\begin{figure}[hbt]
\begin{center}
\[
\begin{tfs}{b}
 	f2: & \begin{tfs}{b}
		f2:	& \tag{1}\begin{tfs}{d}\end{tfs} \\
		f3:	& \tag{1}
	      \end{tfs} \\
	f3: & \begin{tfs}{d}\end{tfs}
\end{tfs}
\hspace{1cm}
\nterm{b(b(\tag{1}d,\tag{1}),d)}
\]
\caption{An example feature structure}
\label{fs-b}
\end{center}
\end{figure}

\section{A TFS Unification Engine}
\label{m0}
\subsection{First-Order Terms vs.\ Feature Structures}
While TFSs resemble FOTs in many aspects, it is important to note the
differences between them. First, TFSs are typed, as opposed to
(ordinary) FOTs.  TFSs are interpreted over more specific domains than
FOTs.  In addition, TFSs label the arcs by feature names, whereas FOTs
use a positional encoding for argument structure. More importantly,
while FOTs are essentially trees, with possibly shared leaves, TFSs
are directed graphs, within which variables can occur anywhere.
Moreover, our system doesn't rule out cyclic structures, so that
infinite terms can be represented, too.  FOTs are consistent only if
they have the same functor and the same arity. TFSs, on the contrary,
can be unified even if their types differ (as long as they have a
non-degenerate least upper bound). Moreover, their arity can differ,
and the arity of the unification result can be greater than that of
any of the unificands.  Consequently, many diversions from the
original WAM were necessary in our design. In the following sections
we try to emphasize the points where such diversions were made.

\subsection{Processing Scheme}
The machine's engine is designed for unifying two TFSs: a {\em
program} and a {\em query}.  The program is compiled once to produce
machine instructions.  Each query is compiled before its execution;
the resulting code is executed prior to the execution of the compiled
program.  Processing a query builds a graph representation of the
query in the machine's memory. The processing of a program produces
code that, during run-time, unifies the program with a query already
resident in memory.  The result of the unification is a new TFS,
represented as a graph in the machine's memory.  In what follows we
interleave the description of the machine, the TFS language it is
designed for and the compilation of programs in this language.

\subsection{Memory Representation of Feature Structures}
Following the WAM, we use a global, one-dimensional array called {\mf
HEAP} of data cells.  A global register {\mf H} points to the
(current) top element of {\mf HEAP}.  Data cells are tagged: STR cells
correspond to nodes, and store their types, while REF cells represent
arcs, and contain the address of their targets. The number of arcs
leaving a node of type $t$ is $Ar(t)$, fixed due to total
well-typedness.  Hence, we can keep the WAM's convention of storing
all the outgoing arcs from a node consecutively following the
node. Given a type $t$ and a feature $f$, the position of the arc
corresponding to $f$ ($f$-arc) in any TFS of type $t$ can be
statically determined; the subgraph that is the value of $f$ can be
accessed in one step.  This is a major difference from the approach
presented in \cite{prl7}, which leads to a more time-efficient system
without harming the elegance of the machine design.

It is important to note that STR cells differ from their WAM analogs
in that they can be dereferenced when a type is becoming more
specific. In such cases, a chain of REF cells leads to the
dereferenced STR cell.  Thus, if a TFS is modified, only its STR cell
has to be changed in order for all pointers to it to `feel' the
modification automatically. The use of self-referential REF cells is
different, too: there are no real (Prolog-like) variables in our
system, and such cells stand for features whose values are temporarily
unknown.

One cell is required for every node and arc, so for representing a
graph of $n$ nodes and $m$ arcs, $n+m$ cells are needed. Of course,
during unification nodes can become more specific and a chain of REF
cells is added to the count, but the length of such a chain is bounded
by the depth of the type hierarchy and path compression during
dereferencing cuts it occasionally. As an example, figure~\ref{heap1}
depicts a possible heap representation of the TFS
\nterm{b(b(\tag{1}d,\tag{1}),d)}.

\begin{figure}[hbt]
\centering
\begin{tabular}{|l|c|c|c|c|c|c|c|c|} \hline
	address: & 1 & 2 & 3 & 4 & 5 & 6 & 7 & 8 \\ \hline
	tag:     & STR&REF&REF&STR&REF&REF&STR&STR\\ \hline
	contents:& b & 4 & 8 & b & 7 & 7 & d & d \\ \hline
\end{tabular}
	\caption{Heap representation of the feature structure
\nterm{b(b(\tag{1}d,\tag{1}),d)}}
        \label{heap1}
\end{figure}

\subsection{Flattening Feature Structures}
Before processing a TFS, its linear representation is transformed to a
set of ``equations'', each having a flat (nesting free) format.  To
facilitate this a set of {\em registers} $\{X_i\}$ that store {\em
addresses} of TFSs in memory is used.  A register {\mf Reg[$j$]} is
associated with each tag $j$ of a normal term.  The flattening
algorithm is straight-forward and similar to the WAM's.
Figure~\ref{eqs} depicts examples of the equations corresponding to
two TFSs.

\begin{figure}[hbt]
\centering
\begin{tabular}{|c|l|}	\hline
Linear representation:	& Set of equations	\\ \hline
\nterm{a(\tag{3}d1,\tag{3})}	& $X1 = a(X2,X2)$	\\
				& $X2 = d1$		\\ \hline
\nterm{b(b(\tag{1}d,\tag{1}),d)}	& $X1 = b(X2,X3)$	\\
				& $X2 = b(X4,X4)$	\\
				& $X4 = d$		\\
				& $X3 = d$		\\ \hline
\end{tabular}
\caption{Feature structures as sets of equations}
\label{eqs}
\end{figure}

\subsection{Processing of a Query}
When processing an equation of the form $X_{i_0} = t(X_{i_1}, X_{i_2},
\ldots)$, representing part of a query, two different instructions are
generated. The first is {\mf put\_node t/n, $X_{i_0}$}, where $n =
Ar(t)$. Then, for every argument $X_{i_j}$, an instruction
of the form {\mf put\_arc $X_{i_0}$, $j$, $X_{i_j}$} is
generated. {\mf put\_node} creates a representation of a node of type
$t$ on top of the heap and stores its address in $X_{i_0}$; it also
increments {\mf H} to leave space for the arcs. {\mf put\_arc} fills
this space with REF cells.

In order for {\mf put\_arc} to operate correctly, the registers it
uses must be initialized. Since only {\mf put\_node} sets the
registers, all {\mf put\_node} instructions must be executed before
any {\mf put\_arc} instruction is. Hence, the machine maintains two
separate streams of instructions, one for {\mf put\_node} and one for
{\mf put\_arc}, and executes the first completely before moving to the
other. This compilation scheme is called for by the cyclic character
of TFSs: as explained in \cite{prl7}, the original single-streamed WAM
scheme would fail on cyclic terms.  Consequently, the order of the
equations becomes irrelevant, and in the actual implementation they
might be processed in any order.

The effect of the two instructions is given in
figure~\ref{put-inst}. Figure \ref{exmpl-code} lists the result of
compiling the term \nterm{b(b(\tag{1}d,\tag{1}),d)}. When this code is
executed (first the {\mf put\_node} instructions, then the {\mf
put\_arc} ones), the resulting representation of the TFS in memory is
the one shown above in figure~\ref{heap1}.

\begin{program}{put-inst}{The implementation of the {\mf put} instructions}{
put\_node t/n,$X_i$ $\equiv$		\\
\label{inst:put-node}
		\> HEAP[H] \get <STR,t>;	\\
		\> $X_i$ \get H;	\\
		\> H \get H + n + 1;	\\
\\
put\_arc $X_i$,offset,$X_j$ $\equiv$		\\
\label{inst:put-arc}
		\> HEAP[$X_i$+offset] \get <REF,$X_j$>;	\\
}
\end{program}
\begin{figure}[hbt]
{\mf
\begin{verbatim}
put_node b/2,X1        % X1 = b(
   put_arc X1,1,X2     %        X2,
   put_arc X1,2,X3     %           X3)
put_node b/2,X2        % X2 = b(
   put_arc X2,1,X4     %        X4,
   put_arc X2,2,X4     %           X4)
put_node d/0,X4        % X4 = d
put_node d/0,X3        % X3 = d
\end{verbatim}
}
\caption{Compiled code for the query \nterm{b(b(\tag{1}d,\tag{1}),d)}}
\label{exmpl-code}
\end{figure}

\subsection{Compilation of the Type Hierarchy}
\label{compile-th}
One of the reasons for the efficiency of our compiler is that it
performs an important part of the unification during compile-time: the
type unification.  The WAM's equivalent of this operation is a simple
functor and arity comparison. It is due to the nature of a typed
system that this check has to be replaced by a more complex
computation.  Efficient methods were suggested for performing
least-upper-bound computation during run time
(see~\cite{lattice-ops}), but clearly computing during compilation
time is preferable.  Since type unification adds information by
returning the features of the unified type, this operation builds new
structures, in our design, that reflect the added knowledge. Moreover,
the WAM's special register S is here replaced by a stack. S is used by
the WAM to point to the next sub-term to be matched against, but in
our design, as the arity of the two terms can differ, there might be a
need to hold the addresses of more than one such sub-term.  These
addresses are stored in the stack.  When the type hierarchy is
processed, the (full) subsumption relation is computed.  Then, a table
is generated which stores, for every two types $t_1,t_2$, the least
upper bound $t = t_1 \unif t_2$.  Moreover, this table lists also the
arity of $t$, its features and their `origin': whether they are
appropriate for $t_1$, $t_2$, both or none of them.  Out of this table
a series of abstract machine language functions are generated. The
functions are arranged as a two-dimensional array called {\mf
unify\_type}, indexed by two types $t_1,t_2$. Each such function
receives one parameter, the address of a TFS on the heap. When
executed, it builds on the heap a skeleton for the unification result:
an STR cell of the type $t_1 \unif t_2$, and a REF cell for each
appropriate feature of it.

Consider {\mf unify\_type[t1,t2](addr)} where {\mf addr} is the
address of a TFS $A$ (of type $t_2$) in memory.  Let $t = t_1 \unif
t_2$, and let $f$ be some feature appropriate for $t$.  If $f$ is
inherited from $t_2$ only, the value of the REF cell is simply set to
point to the $f$-arc in $A$.  If $f$ is inherited from $t_1$ only, a
self-referential REF cell is created. But the information that the
actual value for this cell is yet to be seen must be recorded.  This
is done by means of the global stack S, every element of which is a
pair {\mf <action,addr>}, where {\mf action} is either `copy' or
`unify'. In the case we describe, the action is `copy' and the address
is that of the REF cell. If $f$ is appropriate for both $t_1$ and
$t_2$, a REF cell with the address of the $f$-arc in $A$ is created,
and a `unify' cell is pushed onto the stack. Finally, if $f$ is
introduced by $t$, a VAR cell is created, with $Approp(t,f)$ as its
value. VAR cells are explained in section~\ref{optimization}.  As an
example, we list below (figure~\ref{comp-code}) the resulting code for
the unification the two types {\type a} and {\type b}.

\begin{program}{comp-code}{{\mf unify\_type[a,b]}}{
XXXX \= XXXXXXXXXXXXXXXXXXXXXXXXXXXXXX \= \kill
unify\_type[a,b] (b\_addr) \\
\> build\_str(c);		\> \% since $a \unif b = c$ \\
\> build\_self\_ref\_and\_copy;	\> \% the value of f1 is yet unknown	\\
\> build\_ref(1);		\> \% f2 is the first feature of b	\\
\> build\_ref\_and\_unify(2);	\> \% f3 is the second feature of b	\\
\> 				\> \% but still has to be unified with a\\
\> build\_var(bot);		\> \% f4 is a new structure.	\\
\> bind(b\_addr,H);		\> \% $\equiv$ HEAP[b\_addr] \get <REF,H>\\
\> H \get H + 5;		\\
\> return true;			\\
}
\end{program}

This example code is rather complex; often the code is much simpler:
for example, when $t_2$ is subsumed by $t_1$, nothing has to be
done. As another example, if $t_1$ is subsumed by $t_2$, then
additional features of the program term have to be added to $A$. But
if no such features exist, the only required effect is a change of the
type of $A$.  Another case is when $t_1$ and $t_2$ are
not compatible: {\mf unify\_type[t1,t2]} returns `fail'.
This leads to a call to the function {\mf fail}, which aborts
the unification.

\subsection{Processing of a Program}
The program is stored in a special memory area, the {\mf CODE} area.
Unlike the WAM, in our framework registers that are set by the
execution of a query are not helpful when processing a program. The
reason is that there is no one-to-one correspondence between the
sub-terms of the query and the program, as the arities of the TFSs can
differ.  The registers are used, but (with the exception of $X_1$)
their old values are not retained during execution of the program.

Three kinds of machine instructions are generated when processing a
program equation of the form {$X_{i_0}$ =
t($X_{i_1}$,\ldots,$X_{i_n}$)}. The first instruction is {\mf
get\_structure t/n,$X_{i_0}$}, where $n = Ar(t)$.  For each argument
$X_{i_{j}}$ of $t$ an instruction of the form {\mf unify\_variable
$X_{i_{j}}$} is generated if $X_{i_{j}}$ is first seen; if it was
already seen, {\mf unify\_value $X_{i_{j}}$} is generated.  For
example, the machine code that results from compiling the program
\nterm{a(\tag{3}d1,\tag{3})} is depicted in figure~\ref{program-code}.
The implementation of these three instructions is given in
figure~\ref{unif-insts}.\footnote{ We use the operator `*' to refer to
the contents of an address or a register. }

\begin{figure}[hbt]
{\mf
\begin{verbatim}
get_structure a/2,X1    % X1 = a(
unify_variable X2        %        X2,
unify_value X2           %           X2)
get_structure d1/0,X2    % X2 = d1
\end{verbatim}
}
\caption{Compiled code for the program \nterm{a(\tag{3}d1,\tag{3})}.}
\label{program-code}
\end{figure}

\begin{program}{unif-insts}{Implementation of the get/unify instructions}{
get\_structure t/n,$X_i$ $\equiv$	\\
\label{inst:get-structure}
	\> addr \get deref($X_i$); $X_i$ \get addr;		\\
	\> case HEAP[addr] of		\\
	\>\> <REF,addr>:	\>\>\>\>\>\>\>\> \%
					uninstantiated cell	\\
	\>\>\> HEAP[H] \get <STR,t>;	\\
	\>\>\> bind(addr,H);	\>\>\>\>\>\>\> \%
					HEAP[addr] \get <REF,H> \\
	\>\>\> for j \get 1 to n do HEAP[H+j] \get <REF,H+j>	\\
	\>\>\> for j \get n downto 1 do	push(copy,H+j);	\\
	\>\>\> H \get H + n + 1;	\\
	\>\> <STR,t'>:	\>\>\>\>\>\>\>\> \%
					a node	\\
	\>\>\> if (unify\_type[t,t'](addr) = fail) then fail;	\\
\\
unify\_variable $X_i$ $\equiv$		\\
\label{inst:unify-variable}
	\> <action,addr> \get pop();	\\
	\> $X_i$ \get addr;		\\
\\
unify\_value $X_i$ $\equiv$		\\
\label{inst:unify-value}
	\> <action,addr> \get pop();	\\
	\> case action of		\\
	\>\> copy: HEAP[addr] \get *($X_i$);	\\
	\>\> unify: if (unify(addr,$X_i$) = fail) then fail;		\\
}
\end{program}

The {\mf get\_structure} instruction is generated for a TFS $A_p$ (of
type $t$) which is associated with a register $X_i$. It matches $A_p$
against a TFS $A_q$ that resides in memory using $X_i$ as a pointer to
$A_q$. Since $A_q$ might have undergone some type inference or
previous binding (for example, due to previous unifications caused by
other instructions), the value of $X_i$ must first be
dereferenced. This is done by the function {\mf deref} which follows a
chain of REF cells until it gets to one that does not point to
another, different REF-cell. The address of this cell is the value it
returns.

The dereferenced value of $X_i$, {\mf addr}, can either be a
self-referential REF cell or an STR cell. In the first case, the TFS
has to be built by the program. A new TFS is being built on top of the
heap (using code similar to that of {\mf put\_structure}) with {\mf
addr} set to point to it.  For every feature of this structure, a
`copy' item is pushed onto the stack.  The second case, in which $X_i$
points to an existing TFS of type $t'$, is the more interesting one.
An existing TFS has to be unified with a new one whose type is
$t$. Here the pre-compiled {\mf unify\_type[t,t']} is invoked.

The {\mf unify\_variable} instruction resembles very much its WAM
analog, in the case of {\em read mode}. There is no equivalent of the
WAM's {\em write mode} as there are no real variables in our
system. However, in {\mf unify\_value} there is some similarity to the
WAM's modes, where the `copy' action corresponds to write mode and the
`unify' action to read mode. In this latter case the function {\mf
unify} is called, just like in the WAM.  This function
(figure~\ref{unify-code}) is based upon {\mf unify\_type}.  In
contrast to the latter, the two TFS arguments of {\mf unify} are in
memory, and full unification is performed. The first difference is the
reason for removing an item from the stack S and using it as a part of
the unification process; the second is realized by recursive calls to
{\mf unify} for subgraphs of the unified graphs.

\begin{program}{unify-code}{The code of the {\mf unify} function}{
function unify(addr1,addr2:address): boolean;			\\
\label{fun:unify}
begin								\\
	\> addr1 \get deref(addr1); addr2 \get deref(addr2);	\\
	\> if (addr1 = addr2) then return(true);		\\
	\> if (HEAP[addr1] = <REF,addr1>) then			\\
	\>\> bind(addr1,addr2); return(true);			\\
	\> if (HEAP[addr2] = <REF,addr2>) then			\\
	\>\> bind(addr2,addr1); return(true);			\\
	\> H\_orig \get H; t1 \get *(addr1); t2 \get *(addr2);	\\
	\> if (unify\_type[t1,t2](addr2) = fail) then return (fail);	\\
	\> for i \get 1 to Ar(t1) do				\\
	\>\> <action,addr> \get pop();				\\
	\>\> case action of					\\
	\>\>\> copy: HEAP[addr] \get <REF,addr1+i>;		\\
	\>\>\> unify: if (not (unify (addr,addr1+i))) then return(fail);	\\
	\> bind(addr1,H\_orig);					\\
	\> return(true);					\\
end;								\\
}
\end{program}

When a sequence of instructions that were generated for some TFS is
successfully executed on some query, the result of the unification of
both structures is built on the heap and every register $X_i$ stores
the value of its corresponding node in this graph. The stack S is
empty.

\subsection{Lazy Evaluation of Feature Structures}
\label{optimization}
One of the drawbacks of maintaining total structures is that when two
TFSs are unified, the values of features that are introduced by the
unified type have to be built.  For example, {\mf unify\_type[a,b]}
(figure~\ref{comp-code}) has to build a TFS of type {\type bot}, which
is the value of the $f4$ feature of type {\type c}. This is expensive
in terms of both space and time; the newly built structure might not
be used at all. Therefore, it makes sense to defer it.

To optimize the design in this aspect, a new kind of heap cells,
VAR-cells, is introduced. A VAR cell whose contents is a type {\type
t} stands for the most general TFS of type {\type t}. VAR cells are
generated by the various {\mf unify\_type} functions for introduced
features; they are expanded only when the explicit values of such
features are needed: either during the execution of {\mf
get\_structure}, where the dereferenced value is a VAR cell, or during
{\mf unify}. In both cases the TFS has to be built, by means of
executing the pre-compiled function {\mf build\_most\_general\_fs}
with the contents of the VAR cell as an argument. This function (which
is automatically generated by the type hierarchy compiler) builds a
TFS of the designated type on the heap, with VAR cells instead of REF
cells for the features. These cells will, again, only be expanded when
needed. We thus obtain a lazy evaluation of TFSs that weakly resembles
G\"otz's notion of {\em unfilled feature structures}
(\cite{thilo:master}). Moreover, we gain another important property,
namely that our type hierarchies can now contain loops, since
appropriateness loops can only cause non termination when introduced
features are fully constructed.

\section{Parsing}
\label{parsing}
The previous section delineated a very simple abstract machine,
capable of unifying two simple TFSs. We now add to this machine
control structures that will enable parsing.  We define rules,
grammars and parsing, and then describe how the basic machine is
extended to accommodate the application of a single rule. We sketch
the extensions necessary for manipulating a whole grammar (program).
These extensions were not tested yet.

\subsection{Grammars}
A {\bf multi-rooted structure} (MRS) is a directed, labeled, finite
graph with an ordered non-empty set of distinguished nodes, {\bf
roots}, from which all the nodes are reachable.  A {\bf rule} is a
MRS, where the graph that is reachable from the {\em last} root is the
rule's {\bf head},\footnote{This use of {\em head} must not be confused
with the linguistic one, the core features of a phrase.} and the ones
that are reachable from the rest of the roots form its {\bf
body}.\footnote{Notice that the traditional direction is
reversed. Note also that the head and the body need not be disjoint.}
A MRS is linearly represented as a sequence of terms, separated by
commas, where two occurrences of the same tag, even within two
different terms, denotes reentrancy (that is, the scope of the tags is
the entire sequence of terms). The head is preceded by `$\Rightarrow$'
rather than by a comma.  See~\cite{shuly:tr-lcl-94-8} for the formal
details.

Application of a rule amounts to unifying its body with a MRS resident
in memory and producing its head as a result.  When two TFSs $A_1$ and
$A_2$ are parts of MRSs $\sigma_1$ and $\sigma_2$, respectively, the
{\bf unification} of $A_1$ and $A_2$ {\bf in the context} of
$\sigma_1$ and $\sigma_2$ is defined just like ordinary unification,
but $\sigma_1$ and $\sigma_2$ might be affected by the process.  As an
example, the rule $\rho: \nterm{a(bot,\tag{3}d), d $\Rightarrow$
a(d2,\tag{3})}$ consists of a MRS of length three. When it is applied
to the MRS $\sigma: \nterm{a(d2,d1), d2}$, the result is a new MRS
whose head is \nterm{a(d2,d1)}. $\rho$'s head is modified even though
it does not participate directly in the unification, as it is part of
the context.

A {\bf grammar} is a finite set ${\cal R}$ of rules together with a
{\bf start feature structure} $A_s$.  The lexicon associates with
every word $w$ a TFS $A_w$, its category,\footnote{Words can have more
than one category, but we ignore ambiguity for the sake of
simplicity.} by means of special rules of the form $\langle w
\Rightarrow A_w \rangle$.  The input for the parser, therefore, is a
MRS rather than a string of words.  A MRS $\sigma$ is {\bf derived} by
some TFS $A$ if there exists a rule $\rho \in {\cal R}$ such that $A$
is obtained by unifying $\sigma$ with $\rho$'s body in the context of
$\rho$'s head. We abuse the term `derive' to denote also the reflexive
transitive closure of this relation. $A$ is a {\bf category} if it
derives a substring of some input.\footnote{Categories are TFSs rather
than the atomic symbols of Context Free Grammars.}  The {\bf language}
generated by the grammar is the set of strings of words $\{w_1 \cdots
w_n\}$ such that the category of $w_i$ is $A_i$ for $1 \le i \le n$
and $A_s$ derives $A_1,\ldots,A_n$.

A {\bf dotted rule} (or {\bf edge}) is a MRS that is more specific
than some rule in the grammar, with an additional {\bf dot},
indicating a location preceding some element in the MRS.  An edge is
{\bf complete} if its dot precedes the head and is {\bf active}
otherwise.  We denote dotted rules by $\langle A_1, \ldots, A_i
\edgedot A_{i+1},\ldots, A_n \Rightarrow A_0 \rangle$. Informally, such
a dotted rule asserts that each of $A_1,\ldots,A_i$ derives a string
$\sigma_1,\ldots,\sigma_i$ such that $\sigma_1 \cdots \sigma_i$ is a
substring of the input. $A_{i+1},\ldots,A_n$ still have to derive
$\sigma_{i+1},\ldots,\sigma_n$ in order for $A_0$ to be a category
deriving $\sigma_1 \cdots \sigma_n$.

\subsection{Parsing as Operational Semantics}
\label{parsing-definition}
We view parsing as a computational process along the lines of
\cite{deductive-parsing}.
Given a grammar $(R,A_s)$, an {\bf item} is a triple $[i,\rho,j]$
where $i,j$ are natural numbers and $\rho$ is a dotted rule. A {\bf
state} is a finite set of items.  A computation is triggered by some
input string of words $w_1, \cdots, w_n$ of length $n > 0$.  The
initial state, $\hat{S}$, is $\{[i,\langle w_i \edgedot \Rightarrow
A_i \rangle, i+1] \mid 0 \le i < n\} \cup \{[i,\langle \edgedot \alpha
\Rightarrow B \rangle,i] \mid 0 \le i < n\}$ where $A_i$ is the
category of $w_i$ and $\langle \alpha \Rightarrow B \rangle \in {\cal R}$.
For any state $S$, the next state $S'$ is constructed by the following
transition relation `$\vdash$' (the {\em fundamental rule}):
\begin{quote}
For every $[i,\langle \alpha \edgedot B \beta \Rightarrow A \rangle,
k] \in S$ and $[k,\langle \gamma \edgedot
\Rightarrow B'' \rangle, j] \in S$ such that $B \unif B'' \neq \top$,
add\footnote{If $S \vdash S'$ then $S \subseteq S'$.} $[i,\langle \alpha'
B' \edgedot \beta' \Rightarrow A' \rangle, j]$ to $S'$,
where $\langle \alpha' B' \beta' \Rightarrow A' \rangle$ is obtained by
unifying $B$
and $B''$ in the contexts of $\langle \alpha B \beta A \rangle$ and $\langle
\gamma B'' \rangle$,
respectively.
\end{quote}

A computation is an infinite sequence of states $S_i, i \ge 0$, such
that $\hat{S} = S_0$ and for every $i \ge 0$, $S_i \vdash S_{i+1}$.  A
computation is terminating if there exists some $m \ge 0$ for which
$S_m = S_{m+1}$ (i.e., a fixed-point is reached). A successful
computation is a terminating one, the final state of which contains an
item of the form $[0,\langle \alpha \edgedot \Rightarrow A \rangle,
n]$, where $A_s \subsumes A$; otherwise, the computation fails.
The presence of more than one such item in the final state indicates
that the input can be analyzed in more than one way.

To represent a state of the computation the machine uses a {\bf
chart}, structured as a two-dimensional array storing, in the $(i,j)$
entry, all the dotted rules $\rho$ such that $[i,\rho,j]$ is a member
of the state.  Items are added to the chart by means of an {\bf
agenda} that controls the order of addition.

\subsection{Application of a Single Rule}
\label{m1}
To allow the application of a single rule, the syntax of queries is
extended from simple TFSs to MRSs. The same code is generated for the
queries, with additional {\mf advance} instructions preceding each TFS
of the query.  The {\mf advance} instruction simply increments the
indices of the chart item being manipulated. As a result of executing
the query, the $(i,i+1)$ diagonal of the chart is initialized with
singleton sets of edges.

The syntax of programs is extended, too, from a TFS to a single
MRS. Again, the same code is generated for the TFSs of a program:
program-code for each element of the rule's body and query-code for
the head.  Before the first TFS, a {\mf start\_rule} instruction is
generated. A {\mf move\_dot} and {\mf next\_item} instructions are
generated between two consecutive structures, and after the last one,
the head, an {\mf end\_rule} instruction concludes the generated code.

To understand the effect of these instructions, one must understand
the non-uniform internal representation of dotted rules.  Each such
rule is represented by a record, {\mf edge}, containing three
fields. The {\mf seen} field is a list of pointers to the roots of an
MRS, for the part of the dotted rule preceding the dot. The {\mf
to\_see} field is a pointer to the code area, for the rest of the
rule.  A complete edge is represented as a single TFS, its head, since
the rest of the structures (that are unaccessible from the head) are
irrelevant.  An edge with an initial dot is simply a pointer to code.

Since the rules are applied incrementally, a TFS at a time, care must
be taken of reentrancies. The rules manipulate registers which must
contain the right values when used. To that end the values of the
registers are stored after execution of a part of a rule (that is,
before moving the dot), and the right values are loaded prior to each
such execution (after moving the dot). The field {\mf regs} of
an {\mf edge} stores the saved registers.

{\mf start\_rule} sets the stage for the application of the rule: it
stores the address of the beginning of the query in $X_1$, where {\mf
get\_structure} expects to find it. It also records the values of
$i,j$ and $k$ of the current edges.  {\mf move\_dot} is executed after
the successful unification of one TFS; it copies the newly created
edge, including the values of the registers, to the chart (and
interacts with the agenda).  {\mf next\_item} just restores the
registers' values and resumes execution.  {\mf end\_rule} is executed
once a complete edge is constructed; it adds the edge to the chart and
selects the next edge to work on.

\subsection{Control Structures}
\label{control}
When designing the control module, three parameters
have to be set:
the order of searching chart entries that can combine with a complete
edge $e$;
the order of searching active edges within this chart entry;
and the search strategy: are all the edges that can combine with $e$
computed first, and then their consequences (BFS), or rather all the
consequences of the first such edge, then the next etc.\ (DFS).
The order the chart is searched for active edges is right to left:
from $(i,i)$ to $(0,i)$.  There is no way to decide that a certain
edge in the chosen chart entry is appropriate save by trying to unify
it with the complete edge that was just entered.  Hence all edges in a
chart entry are considered a disjunctive value, and each of them is
tried in turn.  Furthermore, upon initialization each entry on the
diagonal $(i,i)$ of the chart is set to be a disjunction of all the
rules in the grammar.  As for the search strategy, we chose to employ
BFS; some way to record all the edges that were added as consequences
of $e$ is needed, in order to compute their consequences next.

Determining the values of these parameters is program-independent: the
maintenance of the chart is fixed. This fact results from the nature
of the process the machine implements, namely parsing, and has a
desirable consequence: one might change some of these parameters
easily without having to modify the compiler or even the set of
machine instructions. What has to be changed is the data structures
that support the control mechanism.
For lack of space we don't detail the control module. Essentially, it
employs a list of edges, {\mf agenda}, and interacts with the machine
instructions described above through designated functions.

%
%

\section{Conclusion}
\label{conclusion}
As linguistic formalisms become more rigorous, the necessity of well
defined semantics for grammar specifications increases. We presented
an operational semantics for TFS-based formalisms, making use of
an abstract machine specifically tailored for  this kind of
applications. In addition, we described a compiler for a general
TFS-based language. The compiled code, in terms of abstract machine
instructions, can be interpreted and executed on ordinary hardware.
The use of abstract machine techniques is expected to result in highly
efficient processing.

The TFS unification engine and the type hierarchy compiler were
already implemented; the control module will be implemented shortly.
We then plan to enhance the machine by adding specific
values for lists (and perhaps sets). The implementation will
serve as a platform for developing an HPSG grammar for the Hebrew language.

\section*{Acknowledgments}
Part of the work described herein was done while the first author was
visiting the Seminar f\"ur Sprachwiessenschaft in T\"ubingen, Germany.
We wish to thank the Minerva Stipendien Komitee for funding
this visit, and the members of the SFB-340 B4 project
in T\"ubingen, especially Paul King, Thilo G\"otz and John Griffith,
for stimulating discussions.
We also wish to thank Bob Carpenter for his help during this
project, and the anonymous referees for enlightening comments.
This work is supported by a grant from the Israeli Ministry of
Science: ``Programming Languages Induced Computational Linguistics''.
The work of the second author was also partially supported by the Fund
for the Promotion of Research in the Technion.


\end{document}